\documentclass[letterpaper]{article} 
\usepackage{aaai2026}  
\usepackage{times}  
\usepackage{helvet}  
\usepackage{courier}  
\usepackage[hyphens]{url}  
\usepackage{graphicx} 
\usepackage{natbib}  
\usepackage{caption} 
\usepackage{amsmath}    
\usepackage{multirow}   
\usepackage{subcaption}  
\usepackage[dvipsnames]{xcolor}

\usepackage{algorithm}
\usepackage{algorithmic}
\usepackage[capitalize]{cleveref} 
\usepackage{newfloat}
\usepackage{listings}
\DeclareCaptionStyle{ruled}{labelfont=normalfont,labelsep=colon,strut=off} 
\floatstyle{ruled}
\newfloat{listing}{tb}{lst}{}
\floatname{listing}{Listing}
\title{{\boldmath $D^3$-RSMDE: 40$\times$ Faster and High-Fidelity \\ Remote Sensing Monocular Depth Estimation}}
\author{
    Ruizhi Wang\textsuperscript{\rm 1}\equalcontrib,
    Weihan Li\textsuperscript{\rm 1}\equalcontrib,
    Zunlei Feng\textsuperscript{\rm 1,2}, 
    Haofei Zhang\textsuperscript{\rm 2,3} \thanks{Corresponding author},
    Mingli Song\textsuperscript{\rm 1,2,3}, 
    Jiayu Wang\textsuperscript{\rm 4},
    Jie Song\textsuperscript{\rm 1},
    Li Sun\textsuperscript{\rm 5}
}
\affiliations{
    \textsuperscript{\rm 1}School of Software Technology, Zhejiang University, \\ \textsuperscript{\rm 2}State Key Laboratory of Blockchain and Data Security, Zhejiang University, \\ \textsuperscript{\rm 3} Hangzhou High-Tech Zone (Binjiang) Institute of Blockchain and Data Security, \\ \textsuperscript{\rm 4} College of Computer Science and Technology, Zhejiang University, \\ \textsuperscript{\rm 5} Ningbo Global Innovation Center, Zhejiang University
}

\usepackage{bibentry}

\begin{document}

\maketitle

\begin{abstract}
Real-time, high-fidelity monocular depth estimation from remote sensing imagery is crucial for numerous applications, yet existing methods face a stark trade-off between accuracy and efficiency. Although using Vision Transformer (ViT) backbones for dense prediction is fast, they often exhibit poor perceptual quality. Conversely, diffusion models offer high fidelity but at a prohibitive computational cost. To overcome these limitations, we propose \textbf{D}epth \textbf{D}etail \textbf{D}iffusion for \textbf{R}emote \textbf{S}ensing \textbf{M}onocular \textbf{D}epth \textbf{E}stimation ($D^3$-RSMDE), an efficient framework designed to achieve an optimal balance between speed and quality. Our framework first leverages a ViT-based module to rapidly generate a high-quality preliminary depth map construction, which serves as a structural prior, effectively replacing the time-consuming initial structure generation stage of diffusion models. Based on this prior, we propose a \textbf{P}rogressive \textbf{L}inear \textbf{B}lending \textbf{R}efinement (PLBR) strategy, which uses a lightweight U-Net to refine the details in only a few iterations. The entire refinement step operates efficiently in a compact latent space supported by a Variational Autoencoder (VAE). Extensive experiments demonstrate that $D^3$-RSMDE achieves a notable 11.85\% reduction in the Learned Perceptual Image Patch Similarity (LPIPS) perceptual metric over leading models like Marigold, while also achieving over a \textbf{40$\times$ speedup} in inference and maintaining VRAM usage comparable to lightweight ViT models.
\end{abstract}

\section{Introduction}

Real-time, high-fidelity monocular depth estimation from remote sensing images is a fundamental and critical task in computer vision, with profound implications across numerous domains such as autonomous UAV navigation and 3D terrain modeling. One prominent technical approach to this task involves dense prediction architectures employing ViT backbones \cite{ViT}, such as DPT \cite{DPT} and AdaBins \cite{adabins}. Although offering rapid inference, these models have inherent limitations in capturing high-frequency details. Recent studies suggest that ViTs act as low-pass filters, showing a strong tendency to learn global, low-frequency signals while neglecting fine textures \cite{vitnotgood1, vitnotgood2}. This low-pass filtering characteristic of ViTs often leads to perceptually inferior depth maps with blurry details, a deficiency quantifiable by high LPIPS scores \cite{LPIPS}.

\begin{figure}[t]
\centering
\includegraphics[width=\columnwidth]{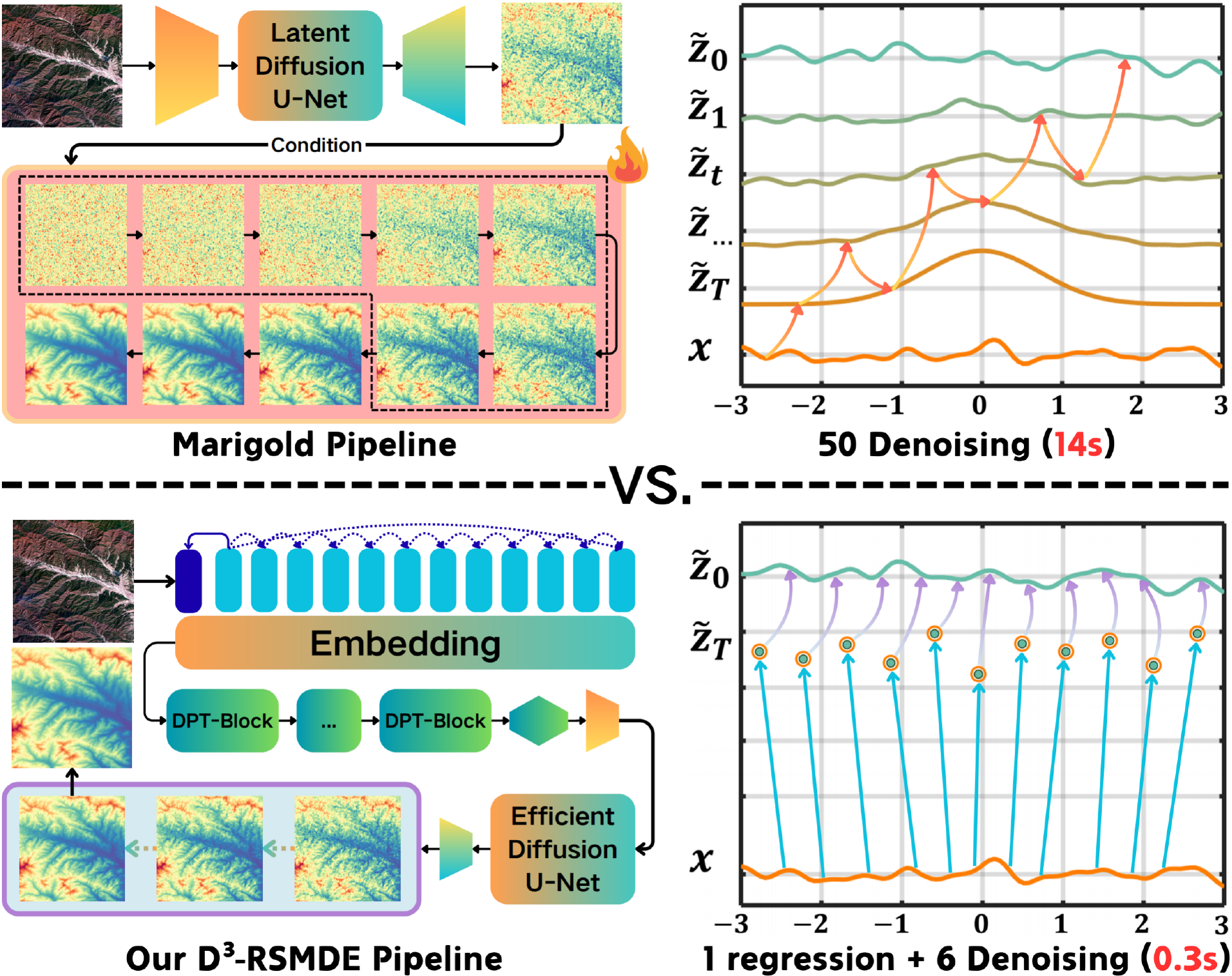} 
\caption{The difference between $D^3$-RSMDE and Marigold. Compared to the multiple denoising reconstructions of  Marigold, our $D^3$-RSMDE firstly adopts efficient ViT to regression coarse depth map and then obtain fine-grained high-fidelity depth map with fewer denoising steps.}
\label{denoise-process} 
\end{figure}

Conversely, an alternative paradigm is diffusion-based generative frameworks, such as Marigold \cite{Marigold} and EcoDepth \cite{ecodepth}. These methods demonstrate remarkable fidelity, generating depth maps with fine-grained textures. This capability is especially useful for remote sensing applications, which often involve intricate surface details. However, their iterative refinement process is computationally intensive and ill-suited for real-time requirements. While conventional acceleration strategies exist, such as optimizing samplers \cite{samplerOptimizer1, samplerOptimizer2, samplerOptimizer3} or employing model distillation \cite{modelDistill1, modelDistill2, modelDistill13}, they typically require the pre-training of a large and resource-intensive base model or sacrifice generative quality for speed \cite{sampler, teacher}, and the limited availability of large-scale training data in the remote sensing domain further constrains the applicability of such distillation-based methods. Furthermore, the iterative nature of diffusion models inherently dedicates the initial, computationally expensive steps to establishing low-frequency macrostructures before refining high-frequency details \cite{diffusionEarly1, diffusionEarly2}. Consequently, these traditional acceleration methods, which speed up the entire process uniformly, fail to radically alter this inefficient workflow.

To investigate this performance bottleneck, we analyzed the depth estimation pipeline of an advanced diffusion model, Marigold, on remote sensing imagery (as illustrated in \cref{denoise-process}). We observed a critical phenomenon: during the entire inference process, which took nearly 14 seconds on an NVIDIA 3090 GPU, a majority of the timesteps (the early stage) were dedicated to establishing the macro-structure and coarse outline of the depth map. In contrast, only a few final steps were used for detail refinement. This insight suggests that the time-consuming initial structure-building phase could be effectively replaced by a more efficient, non-diffusion model, thereby dramatically improving efficiency while preserving high-fidelity details.

Motivated by this observation, we introduce $D^3$-RSMDE, a novel framework designed to achieve a dual optimization of speed and accuracy. First, we leverage a fast ViT-based module, optimized with the Hierarchical Depth Normal (HDN) \cite{HDN} loss function, to efficiently predict a high-quality coarse depth map, thereby completely supplanting the time-consuming initial stage of the diffusion process. Second, we design a lightweight diffusion refinement module that performs coarse-to-fine detail enhancement over a significantly shorter trajectory. The core of this module is our innovative PLBR strategy, which ensures both accuracy and efficiency. In PLBR, at each refinement step, the model is conditioned on both the original coarse map and the output from the previous step, with their influences dynamically attenuated. This provides a stable global structure reference while preventing excessive interference with detail synthesis, enabling controllable and precise reconstruction. Finally, to further accelerate this process for large-scale remote sensing images, we incorporate a Variational Autoencoder (VAE) \cite{VAE}, mapping the entire refinement operation into a compact latent space to drastically reduce computational overhead.

In summary, our main contributions are as follows: 
\begin{itemize} 
    \item We propose $D^3$-RSMDE, specifically designed for efficient and high-fidelity monocular depth estimation from remote sensing imagery, obtaining over 40$\times$ speedups compared to Marigold.
    \item We introduce an innovative PLBR and leverage a VAE to operate in the latent space, significantly enhancing accuracy and computational efficiency.
    \item Through extensive experiments on five datasets, we demonstrate that our method achieves SOTA or second-best performance, while its efficiency is comparable to that of lightweight ViT-based models, effectively resolving the bottlenecks of existing technologies.
\end{itemize}

\section{Related Works}

\subsection{ViT-based Monocular Depth Estimation}

ViTs \cite{ViT} have been widely adopted for MDE, owing to their powerful global feature extraction capabilities. This line of research has produced a series of models prioritizing efficiency and global consistency.

One of the early explorations in this domain was AdaBins \shortcite{adabins}. This work uses Transformer \cite{Transformer} to reformulate depth regression as a classification problem and significantly improved accuracy. Subsequently, DPT \shortcite{DPT} pioneered the use of the ViT architecture as an encoder backbone, demonstrating its superiority over traditional CNNs in capturing global image context and laying a solid foundation for subsequent research.

To enhance model generalization across diverse scenes, Eftekhar et al. introduced Omnidata \shortcite{omnidata}, an innovative parametric pipeline for sampling and rendering multi-task vision datasets from real-world 3D scans. Subsequent studies further proposed the HDN loss function \shortcite{HDN}, which enhances geometric consistency by enforcing constraints on surface normals at multiple scales. These methods collectively form the foundation for the first component of our work: the rapid generation of a structural depth map prior.

More recently, Depth Anything \cite{depthAnything}, demonstrates exceptional zero-shot capabilities on general-purpose scenes. However, its performance fails to generalize to the domain of remote sensing imagery. The powerful depth priors learned from near-view perspectives are ill-suited for the unique top-down viewpoints, distinct geometric properties, and the absence of conventional depth cues found in remote sensing data. This domain gap underscores the necessity of developing specialized algorithms tailored for the unique challenges of monocular depth estimation in the remote sensing context.

\subsection{Diffusion-based Monocular Depth Estimation} 

Diffusion models \cite{diffusion} have opened new frontiers in high-quality depth map synthesis, with their core strength lying in unparalleled detail recovery and realism. ECoDepth \shortcite{ecodepth} fuses pretrained ViT embeddings with a diffusion model for monocular depth estimation: global semantic features are first extracted via ViT, then conditioned alongside the input image in the diffusion denoising process to produce depth maps, achieving outstanding results on near-view scenes.

At the same time, Marigold \shortcite{Marigold} ingeniously repurposes a pre-trained text-to-image diffusion model (like Stable Diffusion) for zero-shot monocular depth estimation. By incorporating depth information into Gaussian noise, it can generate depth maps with remarkable realism and fine detail. These works demonstrate the potent potential of diffusion models for detail generation, they also highlight their inherent drawback of high computational cost, a core problem our work aims to solve.

 \begin{figure*}[t] 
\centering
\includegraphics[width=\textwidth]{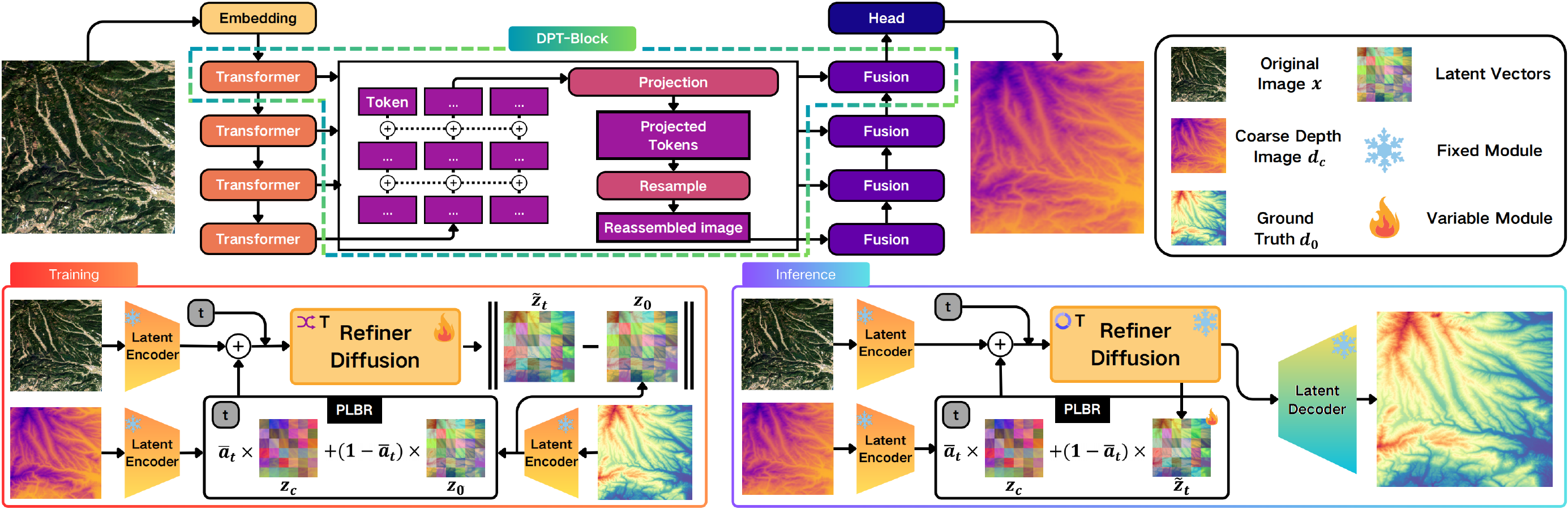} 
\caption{\textbf{The framework of our $D^3$-RSMDE.} During the training process, ViT first performs regression on the input original remote sensing images \boldmath{$x$} to obtain the coarse depth map construction \boldmath{$d_c$}, and then together with Ground Truth \boldmath{$d_0$} and the \boldmath{$x$}, obtains the samples for training Refiner Diffusion through \textbf{PLBR}. In the inference process, the \boldmath{$d_0$} is replaced by the output of each step of Refiner Diffusion to obtain a refined and high-fidelity remote sensing depth estimation map.}
\label{fig:architecture}
\end{figure*}

Leveraging diffusion models for refinement is a nascent trend in computer vision. However, existing approaches are largely ill-suited for efficient, continuous depth map refinement: DDRM \cite{DDRM} in image restoration aims to reverse physical degradation, not correct neural network prediction ambiguity; SegRefiner \cite{SegRefiner} in segmentation tackles discrete labels instead of continuous values; DifFlow3D \cite{liu2024difflow3d} predicts scene flow through DDIM-based iterative diffusion and achieves strong performance, yet it is designed for irregular point clouds and cannot be directly applied to monocular depth estimation in remote sensing; and DCTPose \cite{DCTPose} in pose estimation refines sparse coordinates rather than depth maps. These fundamental differences in task definition and data structure highlight a critical gap in monocular depth estimation for a framework designed specifically for efficient, high-fidelity refinement of fine-grained remote-sensing depth maps, our proposed $D^3$-RSMDE is precisely intended to fill this gap.

\subsection{VAE for Diffusion Models}
To mitigate the immense computational expense of operating directly in the high-dimensional pixel space, researchers have shifted to performing the diffusion process within a compact latent space. The theoretical foundation for this paradigm was laid by Kingma \& Welling \cite{VAE} with the introduction of the VAE, which pioneered the use of variational Bayesian methods to learn a mapping from data to a probabilistic latent space.

Years later, Rombach et al. \cite{stablediffusion} applied this concept to generative models, proposing Latent Diffusion Models and designing the crucial AutoencoderKL (AEKL). This VAE, widely adopted by models like Stable Diffusion, efficiently compresses images into a low-dimensional latent space, drastically reducing the computational cost of the diffusion process and establishing it as a mainstream approach.

However, the standard AEKL faces an optimization dilemma between ``reconstruction" and ``generation": it must both ensure high-fidelity image reconstruction and provide a smooth, regularized latent space for the diffusion U-Net. To address this, various improvements have been proposed. For example, VA\_VAE \cite{VAVAE}  demonstrates that by introducing a lightweight auxiliary decoder to exclusively handle the reconstruction loss, the primary decoder can be freed to focus on generative quality. This decoupled design, while retaining the AEKL backbone, significantly accelerates training and enhances final generation quality. The development of these VAE technologies is the key underpinning that enables our design of a computationally feasible, lightweight diffusion refinement module.

\section{Method}

\subsection{Overview} 

The $D^3$-RSMDE framework is a hybrid architecture that efficiently combines different paradigms for accurate monocular depth estimation from remote sensing images. It first employs a ViT-based module to quickly generate a structurally consistent coarse depth map, avoiding the slow contour construction of traditional diffusion methods. A lightweight diffusion module then refines this depth scene in a few steps in a compact latent space, producing a detailed depth output.

\subsection{Preliminary Scene Structuring}

The preliminary depth scene estimation module is designed to produce a globally consistent and structurally coherent initial depth map scene for subsequent refinement. It refers to the DPT model and employs a hybrid architecture that combines a ViT encoder with a convolution-based decoder. The encoder divides the input image into non-overlapping patches of size $p \times p$, yielding $N_p = \frac{HW}{p^2}$ flattened tokens. Each token is linearly projected into a $D$-dimensional embedding space, augmented with learnable positional encodings and a global readout token. These tokens $\{t_0, t_1, \dots, t_{N_p}\}$ are processed by $L$ layers of multi-head self-attention, excelling at long-range dependency modeling to ensure global structural consistency.

The $N_p$ tokens are then reassembled into a feature map of shape $\frac{H}{p} \times \frac{W}{p} \times D$ based on their original patch positions. A Resample layer adjusts the resolution using a $1 \times 1$ convolution to map channel dimension to $\hat{D}$, followed by either a strided $3 \times 3$ convolution for downsampling (if $s \geq p$) or a transposed $3 \times 3$ convolution with stride $p/s$ for upsampling (if $s < p$). This reassembly and resampling process is performed at transformer layers $\{3,6,9,12\}$ to extract multi-scale representations. These are subsequently fused in a top-down manner using a RefineNet-style decoder, where each stage doubles the spatial resolution and merges features hierarchically. The final feature map, at half the original input resolution, is passed to a task-specific output head to generate the coarse depth map.

To supervise the model, we employ the HDN loss, which balances global structural consistency with local detail preservation. The HDN loss is defined as:
$$L_{\mathrm{HDN}} 
=\frac{1}{M}\sum_{i=1}^M\Bigl(\frac{1}{|\mathcal U_i|}\sum_{u\in\mathcal U_i}\bigl|N_{u_i}(d_i)-N_{u_i}(\tilde{d}_i)\bigr|\Bigr),$$
where the normalized representations is given by:
{$$\mathcal{N}_{u_i}(d_i)=\frac{d_i-\mathrm{median}_{u_i}(\mathbf{d})}{\frac{1}{|u_i|}\sum_{j=1}^{|u_i|}|d_i-\mathrm{median}_{u_i}(\mathbf{d})|},$$} 
$\mathrm{median}_{u_i}$ computes the median depth of locations, $M$ is the total number of effective pixels, $i$ represents each pixel, $\mathcal{U}_i$ denotes the set of multi-scale contexts to which pixel $i$ belongs, $d$ is the ground truth and $\tilde{d}$ is the predicted depth map. These contexts are constructed using three strategies: spatial grid partitioning, depth-range segmentation, and depth-quantile grouping. For each context, a shared SSI module \cite{HDN} computes the MAE, and the errors are aggregated across contexts and scales.

\subsection{Progressive Detail Refinement}

Although the coarse depth map generated by the initial prediction module exhibits a low MAE, it scores poorly on LPIPS, appearing visually blurry and lacking in high-frequency details. To address this, we designed a Dynamic Guided Diffusion Refiner. This design departs from the conventional paradigm of Markovian based diffusion models \cite{Markov} that reconstruct data from pure noise. Inspired by SegRefiner \cite{SegRefiner}, we instead formulate a non-Markovian based coarse-to-fine refinement process. This mechanism ensures that the globally consistent structural information from the coarse depth map continuously guides the entire refinement process. This module greatly shortens the process of diffusion to reconstruct depth map details from pure noise, which can efficiently recover clear and realistic depth details with only a few iterations.

\subsubsection{Efficient Diffusion backbone.}

The core is a conditional diffusion denoising model $f$, tasked with predicting the latent representation of the refined depth map at a given timestep $t$. To strike a balance between computational efficiency and expressive power, our model operates entirely within a compact latent space defined by a pre-trained VAE.

\begin{table*}[t]
\footnotesize
\centering

\renewcommand{\arraystretch}{1.5}

\setlength{\tabcolsep}{0.6mm}{

\begin{tabular}{c|ccccc|ccccc|ccccc|ccccc}
\hline

\multirow{2}{*}{\textbf{Model}} & \multicolumn{5}{c|}{\textbf{MAE} $\downarrow$} & \multicolumn{5}{c|}{$\boldsymbol{\delta^3} \uparrow$} & \multicolumn{5}{c|}{\textbf{PSNR} $\uparrow$} & \multicolumn{5}{c}{\textbf{LPIPS} $\downarrow$} \\ \cline{2-21} %
& J\&K & SA & Med & Swi & Ast & J\&K & SA & Med & Swi & Ast & J\&K & SA & Med & Swi & Ast & J\&K & SA & Med & Swi & Ast \\
\hline
Adabins & 16.7 & 28.4 & 29.0 & 19.6 & 44.3 & 79.9 & 68.9 & \underline{86.3} & 93.1 & \underline{73.2} & 22.3 & 17.4 & 17.7 & 21.2 & 14.7 & 0.181 & 0.405 & 0.367 & 0.127 & 0.528 \\
DPT & 17.3 & 34.2 & 29.7 & 30.9 & 43.5 & 77.3 & 62.5 & 81.6 & 84.6 & 72.7 & 22.2 & 16.7 & 17.8 & 17.6 & 14.9 & 0.313 & 0.604 & 0.520 & 0.204 & 0.579 \\
Omnidata & 20.1 & 30.7 & 28.0 & 19.2 & 42.6 & 61.9 & 67.8 & 80.9 & 90.8 & 72.1 & 21.2 & 18.5 & 18.2 & 21.6 & 15.0 & 0.354 & 0.482 & 0.479 & 0.135 & 0.553 \\
\hline
Pix2pix & 24.5 & 39.3 & 39.4 & 38.9 & 44.3 & 68.9 & 55.1 & 72.0 & 76.8 & 69.3 & 18.6 & 15.2 & 15.1 & 15.5 & 14.3 & 0.450 & 0.485 & 0.434 & 0.775 & 0.937 \\
\hline
Marigold & 14.2 & 23.7 & 24.7 & 21.3 & \underline{40.0} & \underline{83.1} & \underline{71.7} & 85.8 & 89.6 & 72.8 & \underline{24.3} & 19.6 & 19.3 & 21.4 & \underline{15.7} & \textbf{0.162} & \underline{0.326} & 0.329 & 0.144 & \textbf{0.488} \\
EcoDepth & 26.4 & 49.0 & 49.0 & 37.4 & 43.3 & 65.7 & 49.4 & 67.4 & 77.9 & 69.9 & 17.9 & 13.3 & 13.2 & 16.0 & 14.5 & 0.461 & 0.428 & 0.563 & 0.265 & 0.702 \\
\cline{1-1} \cline{2-21} 

\textbf{\boldmath$D^3$-RSMDE (VA\_VAE)} & \underline{13.6} & \underline{21.7} & \underline{23.4} & \underline{14.1} & 41.7 & 79.1 & 70.0 & 85.9 & \underline{93.3} & 67.8 & 23.7 & \underline{20.1} & \underline{20.0} & \underline{24.2} & 15.1 & 0.203 & 0.366 & \underline{0.301} & \underline{0.107} & 0.574 \\
\textbf{\boldmath$D^3$-RSMDE (AEKL)} & \textbf{12.7} & \textbf{20.5} & \textbf{22.1} & \textbf{13.4} & \textbf{36.1} & \textbf{83.3} & \textbf{73.9} & \textbf{88.1} & \textbf{94.3} & \textbf{76.1} & \textbf{24.5} & \textbf{20.6} & \textbf{20.4} & \textbf{24.8} & \textbf{16.2} & \underline{0.180} & \textbf{0.318} & \textbf{0.290} & \textbf{0.104} & \underline{0.511} \\
\hline
\end{tabular}
}
\caption{Quantitative analysis of SOTA methods. The best result is highlighted in \textbf{bold} and the second best result is \underline{underlined}.}
\label{tab:my_quantitative_results}
\end{table*}

Compared to a conventional Stable Diffusion U-Net, our model excises modules that are superfluous for our depth map refinement task, such as text cross-attention and multi-source conditioning frameworks. This results in a significantly more lightweight architecture specialized for refining depth details from image and timestep information. The detailed Unet architecture can be found in Appendix A.4.

\subsubsection{Progressive Linear Blending Refinement.} 

Unlike traditional diffusion models that gradually transform data into pure Gaussian noise, we introduces a refiner-specific strategy called PLBR. This process is different from the traditional diffusion strategy based on Markov. PLBR is based on a non-Markovian process, linearly interpolates between the high-quality ground truth depth map and a coarse depth map during training. During inference, this process is reversed through a Progressive Refinement procedure.

The goal of the forward process is to generate training samples of varying ``levels of noise" for the model $f$. We define two key inputs: the ground truth depth map $d_0$ and a coarse depth map $d_c$ generated by a DPT module. These inputs are first encoded by the VAE into their respective latent representations, $z_0$ and $z_c$.

We design a diffusion schedule coefficient: $$\bar{\alpha}_t=\frac{\epsilon}{T-1}(T-t-1),$$ where $t\in[0,...,T-1]$, $\epsilon$ is a positive constant that close to 1 but not equal to 1, ensuring that the coarse depth map's contribution is always present (In our experiment, $\epsilon=0.8$). The blended latent representation $z_t$ at any timestep $t$ is generated via the following linear interpolation formula:
$$z_t = \bar{\alpha}_t z_0 + (1 - \bar{\alpha}_t) z_c,$$

This process simulates a continuous transition from fine to coarse. When t is small, $\bar{\alpha}_t$ is close to 1, and the primary component of $z_t$ is the ground truth $z_0$ . As $t$ increases, $\bar{\alpha}_t$ decreases, and $z_t$ progressively approaches the coarse representation $z_c$ . At the same time, the original remote sensing image $x$ will also be used as the input of additional information concat into the model to provide more basic information. This strategy enables the model to learn how to recover fine depth structures from inputs of varying coarseness across the entire refinement trajectory.

The inference is an iterative refinement process that aims to progressively recover a high-quality depth map $d_0$ , starting from the coarse map $d_c$ . The process begins at $t=T-1$ and proceeds backward to $t=0$.

\begin{itemize}
\item First, we encode the initial coarse depth map $d_c$ into its latent representation $z_c$ and set it as the input for the first timestep, i.e., $z_{T-1}=z_c$ .

\item At each timestep $t$, the $f$ model receives the current input $z_t$ , the remote sensing image latent $z_{x}$, and the timestep embedding $e_t$ , and predicts the final refined latent representation $\tilde{z}_{0|t}$:
$$\tilde{z}_{0|t}=f([z_{x},z_t],e_t),$$

\item Then, we employ a novel progressive refinement strategy to generate the input for the next step $z_{t-1}$. Unlike conventional DDPMs, which rely on the previous state $z_t$ and the current prediction $\tilde{z}_{0|t}$ to estimate $z_{t-1}$, our method directly blends the new prediction with the original coarse representation $z_c$. This ensures that each refinement step is anchored to the initial coarse structure, preventing error accumulation, follows the update rule: 
$$z_{t-1}=\bar{\alpha}_{t-1}\tilde{z}_{0|t}+(1-\bar{\alpha}_{t-1})z_{c},$$

\end{itemize}

This iterative process continues until $t=0$, yielding the final latent representation $\tilde{z}_{0|1}$. Finally, the VAE decoder transforms $\tilde{z}_{0|1}$ back into the pixel space to produce the final, refined depth map $\tilde{d}_0$.

\section{Experiment}

\subsection{Experimental Settings} 
\subsubsection{Benchmark and Metrics.}
In order to comprehensively evaluate the performance of the model in different terrain, resolution, and dataset sizes, we chose 5 datasets from RS3DBench\cite{wang2025rs3dbench}: Japan + Korea (2,650 pairs, coastal mountainous terrain, 30 m resolution, J\&K), Southeast Asia (7,000 pairs, plains and hills, 30 m resolution, SA), Mediterranean (29,225 pairs, desert and plateau, 30 m resolution, Med), Australia (1,249 pairs, plain, 5m resolution, Ast), Switzerland (4,827 pairs, mountain, 2m resolution, Swi). In order to better reflect human perception of depth map quality, in addition to traditional MAE, $\delta^3$ and PSNR, etc., we also introduce LPIPS metric. LPIPS is a trained perceptual loss metric that uses the pre-trained AlexNet network to calculate the structural and texture similarity between images to evaluate the perceptual effect of image reconstruction or generation quality. A detailed description of the metrics can be found in Appendix A.1.

\begin{table*}[t]
\centering

\renewcommand{\arraystretch}{1.3}

\resizebox{\textwidth}{!}{
\begin{tabular}{l|ccc|ccc|ccc|ccc|ccc}
\hline
\multirow{2}{*}{\textbf{Configuration}} & \multicolumn{3}{c|}{\textbf{MAE} $\downarrow$} & \multicolumn{3}{c|}{\textbf{RMSE} $\downarrow$} & \multicolumn{3}{c|}{$\boldsymbol{\delta^3} \uparrow$} & \multicolumn{3}{c|}{\textbf{PSNR} $\uparrow$} & \multicolumn{3}{c}{\textbf{LPIPS} $\downarrow$} \\ \cline{2-16}
& J\&K & SA & Med & J\&K & SA & Med & J\&K & SA & Med & J\&K & SA & Med & J\&K & SA & Med \\
\hline
DPT (Baseline) & 17.3 & 34.2 & 29.7 & 22.1 & 42.3 & 36.4 & 77.3 & 62.5 & 81.6 & 22.2 & 16.7 & 17.8 & 0.313 & 0.604 & 0.520 \\
ViT Module Only & 12.8 & 23.3 & 23.2 & 17.2 & 30.4 & 28.5 & 79.2 & 70.3 & 84.8 & \underline{24.4} & 19.8 & 20.2 & 0.305 & 0.503 & 0.439 \\
\hline
Full Model (w/o VAE, T=3) & 13.0 & 22.1 & 22.9 & 17.2 & 28.6 & 28.0 & 83.1 & 72.2 & 86.2 & 24.3 & 20.1 & \underline{20.3} & 0.222 & 0.361 & 0.335 \\
Full Model (w/o VAE, T=6) & \textbf{12.4} & \underline{21.6} & \underline{22.6} & \textbf{16.7} & \underline{27.9} & \underline{27.5} & \textbf{84.9} & \underline{72.4} & \underline{86.6} & \textbf{24.5} & \underline{20.3} & \textbf{20.4} & 0.218 & \underline{0.343} & 0.344 \\
Full Model (w/o VAE, T=10) & 13.5 & 23.0 & 24.5 & 17.6 & 29.2 & 29.5 & 82.5 & 69.3 & 85.0 & 24.1 & 19.8 & 19.9 & 0.239 & 0.365 & 0.365 \\
\hline
Full Model (AEKL, T=3) & 13.3 & 21.8 & 24.3 & 17.4 & 28.3 & 29.8 & 82.3 & 71.5 & 83.0 & 24.2 & 20.1 & 19.7 & 0.199 & 0.350 & 0.323 \\
Full Model (AEKL, T=6) & \underline{12.7} & \textbf{20.5} & \textbf{22.1} & \underline{16.9} & \textbf{26.6} & \textbf{27.1} & \underline{83.3} & \textbf{73.9} & \textbf{88.1} & \textbf{24.5} & \textbf{20.6} & \textbf{20.4} & \textbf{0.180} & \textbf{0.318} & \textbf{0.290} \\
Full Model (AEKL, T=10) & 13.3 & 21.9 & 24.1 & 17.3 & 28.3 & 29.7 & 80.8 & 72.0 & 84.5 & 24.1 & 20.0 & 19.6 & \underline{0.187} & 0.349 & 0.334 \\
\hline 
Full Model (VA\_VAE, T=3) & 14.4 & 21.9 & 24.3 & 19.3 & 28.3 & 29.8 & 78.2 & 69.0 & 83.0 & 23.2 & 20.1 & 19.7 & 0.260 & 0.388 & 0.323 \\
Full Model (VA\_VAE, T=6) & 13.6 & 21.7 & 23.4 & 18.2 & 28.1 & 28.8 & 79.1 & 70.0 & 85.9 & 23.7 & 20.1 & 20.0 & 0.203 & 0.366 & \underline{0.301} \\
Full Model (VA\_VAE, T=10) & 14.9 & 21.9 & 24.1 & 20.1 & 28.6 & 29.7 & 77.8 & 70.4 & 84.5 & 22.7 & 20.0 & 19.6 & 0.324 & 0.374 & 0.334 \\
\hline
\end{tabular}
}
\caption{Ablation study of $D^3$-RSMDE. The best result is highlighted in \textbf{bold} and the second best result is \underline{underlined}.}
\label{ablation_study}
\end{table*}

\subsubsection{Implementation Details.} We release two model versions using AEKL \shortcite{stablediffusion} and VA\_VAE \shortcite{VAVAE}, respectively. For reproducibility, the random seed for all experiments was fixed to 42. The initial prediction module was trained using the HDN loss with an initial learning rate (LR) of $5\times10^{-5}$ and a weight decay of $10^{-4}$ . We employed a LR scheduler that reduces the LR by a factor of $0.6$ upon a validation loss plateau of $5$ epochs. To generate unbiased training and testing sets for the diffusion refiner, we performed 5-fold cross-validation on the outputs of the initial prediction module. The refiner was subsequently trained on these outputs using an L1 loss. The initial LR was set to $1\times10^{-4}$ with $T=6$, $\epsilon=0.8$ and a similar LR scheduler (patience=$5$, factor=$0.5$). Additional hyperparameter details are provided in the Appendix A.2.

\subsubsection{Baselines for Comparison}
To comprehensively benchmark our $D^3$-RSMDE model against existing technologies, we selected a series of baseline models that have achieved SOTA performance and are based on different mainstream architectural paradigms.
\begin{itemize}
\item \textbf{ViT Models}: We include DPT, Omnidata and AdaBins. These three models are representative works that leverage ViT for efficient monocular depth estimation.
\item \textbf{Diffusion Models}: We compare against Marigold and EcoDepth. They represent the cutting edge in high-fidelity depth synthesis using diffusion processes.
\item \textbf{GAN-based Model}: We also include Pix2pix \cite{pix2pix}, a GAN-based model specifically optimized for remote sensing monocular depth estimation.
\end{itemize}

\subsection{Quantitative Analysis}

\begin{figure*}[h]
\centering
\includegraphics[width=\textwidth]{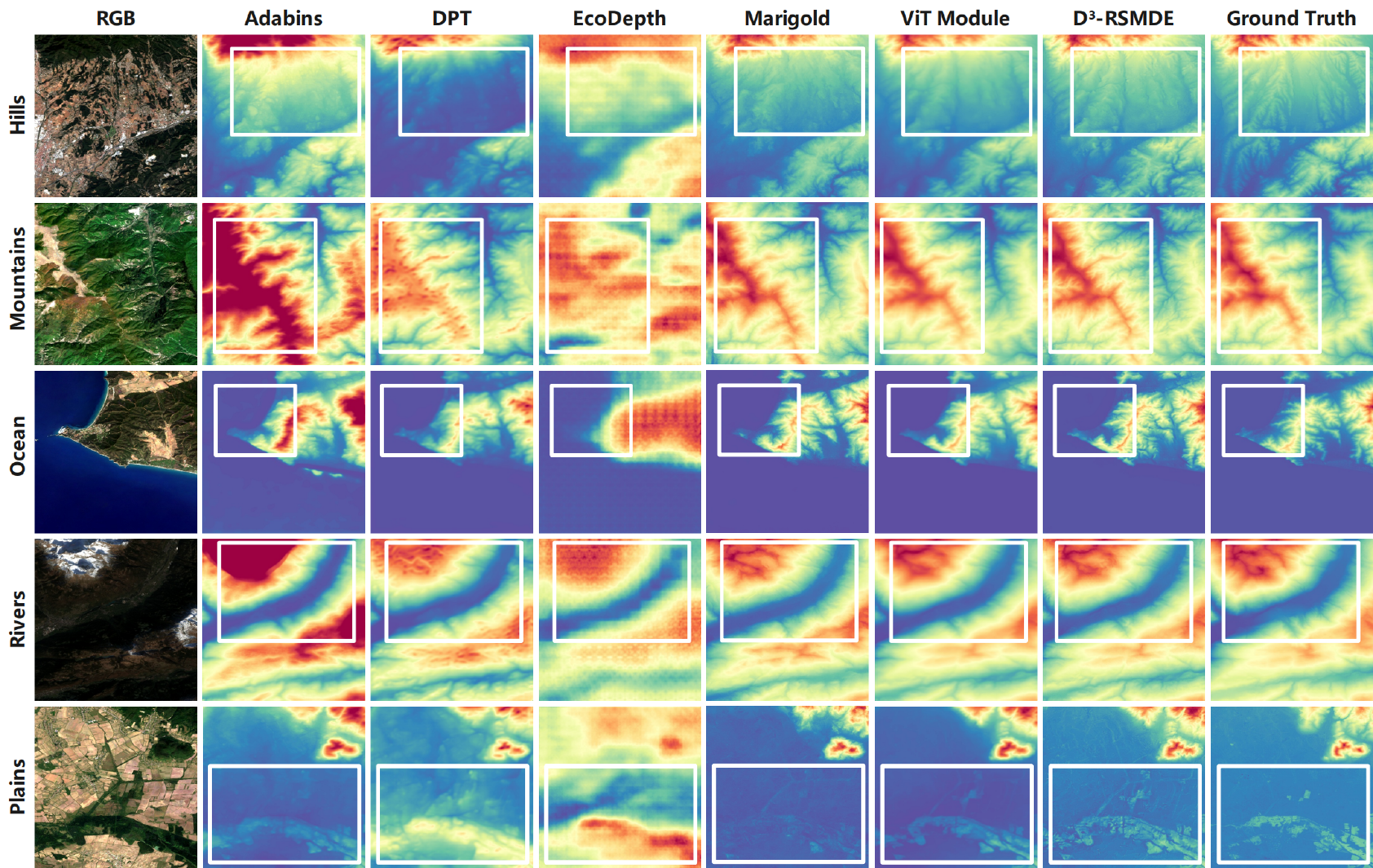} 
\caption{Comparison of our $D^3$-RSMDE and some SOTA methods in different categories of remote sensing images.}
\label{compare7}
\end{figure*}

The quantitative evaluation results for all models are summarized in \cref{tab:my_quantitative_results}, with visual comparisons against representative models shown in \cref{compare7}. The results comprehensively demonstrate that our proposed $D^3$-RSMDE framework achieves SOTA or second-best performance across the majority of metrics on all test datasets. Our framework significantly outperforms the GAN-based Pix2pix and the ViT-based Adabins, DPT, and Omnidata. More remarkably, even when compared against retrained Marigold, a model renowned for its high-fidelity synthesis, our $D^3$-RSMDE achieves a substantial relative improvement of up to 13.50\% in MAE and 11.85\% in LPIPS, showcasing its exceptional overall performance. The general observation is that all diffusion-based models, except the retrained EcoDepth, outperform the LPIPS perception metric. This highlights the distinct advantage of the diffusion architecture for generating photorealistic RSMDE. A detailed diagnostic analysis for the underperformance of EcoDepth, which is also a diffusion model, is provided in Appendix A.3.

\begin{figure}[t]
    \centering 
    \includegraphics[width=\columnwidth]{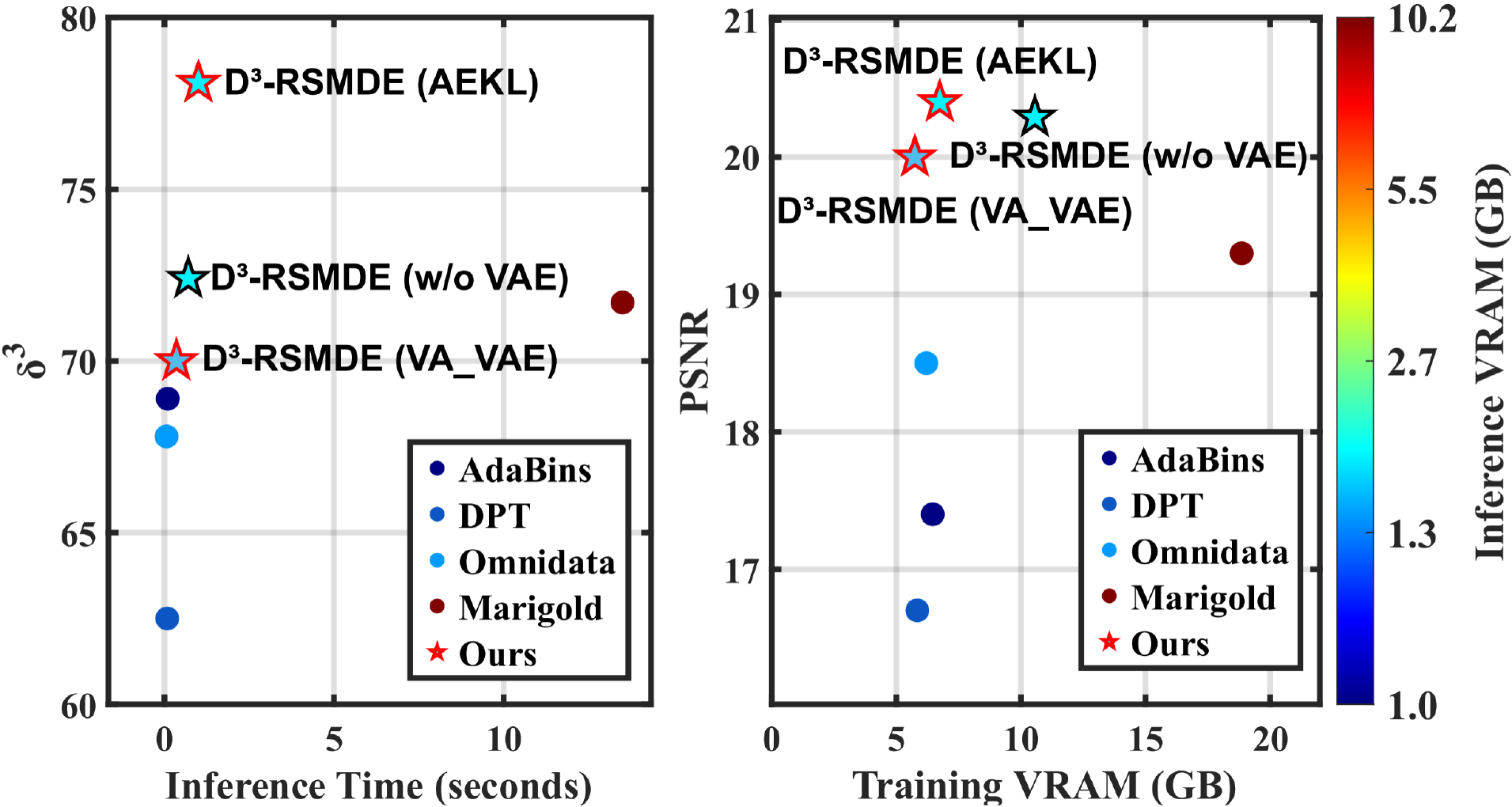}
    \caption{Comparison of model efficiency.} 
    \label{fig:compare_efficiency}
\end{figure}

\subsection{Efficiency Analysis}

To comprehensively evaluate the computational efficiency of our $D^3$-RSMDE framework, we conducted a systematic analysis of the average inference time per image, training time per epoch while training Japan + Korea Dataset, maximum inference VRAM usage, and maximum training VRAM usage (with a batch size of 2). All experiments were performed on a consistent hardware/software platform (Ubuntu 16.04, Intel CPU E5-2699 v4, NVIDIA 3090 GPU, 125G memory, python3.10.6), with the results presented in \cref{fig:compare_efficiency}, where $D^3$-RSMDE shows its diffusion module.

Experimental data reveals that our framework demonstrates a decisive advantage across all efficiency metrics when compared to SOTA diffusion methods like Marigold and EcoDepth. Most notably, $D^3$-RSMDE achieves an inference speed over \textbf{40 times} faster than Marigold while also significantly reducing training time. In terms of resource consumption, both the training and inference memory footprints of our model are substantially lower than other diffusion models. Furthermore, it is noteworthy that the VRAM of our model is on par with that of lightweight ViT-based model such as DPT and Omnidata at inference and training time. This result provides strong evidence that $D^3$-RSMDE successfully reduces computational overhead to a level comparable to non-generative models, all while retaining the high-quality synthesis capabilities of diffusion models.

\subsection{Ablation Study}

In this section, we conduct a series of detailed ablation studies to individually validate the effectiveness and efficiency of the key components within our $D^3$-RSMDE framework. The comprehensive results are summarized in \cref{ablation_study}.

\subsubsection{Effectiveness of the ViT Module.}

First, we compared the performance of our ViT Module against a standard DPT. As shown in \cref{ablation_study}, although our module is based on the DPT architecture, its optimization with the HDN loss function leads to a significant improvement in the quality of the initial depth map. This provides a superior structural prior for the subsequent diffusion refinement, thereby effectively enhancing the final output accuracy.

\subsubsection{Efficiency of using VAE.}

Next, we evaluated the role of performing refinement in the latent space using a VAE. The results indicate that, with the same number of denoising steps, the model variant with a VAE achieves comparable accuracy to a variant that performs refinement directly in the pixel space. However, considering the efficiency data from \cref{fig:compare_efficiency}, we found that incorporating a VAE improves training speed by 54.91\% and reduces training VRAM by 36.17\%. This provides strong evidence that latent space diffusion can dramatically enhance training efficiency and lower the resource threshold without compromising model performance.

\subsubsection{Impact of Denoising Steps.}

We investigated the impact of varying the number of denoising steps (T) on the final result. As shown in \cref{ablation_study}, the vast majority of those achieving the SOTA metric are concentrated in the models with step=6 without VAE and with AEKL. The performance of the model improves markedly as the number of steps increases from 3 to 6, suggesting that T=3 allows for an insufficient refinement process that does not fully leverage the model's detail recovery capabilities. However, when the steps are further increased to 10, performance slightly degrades. We hypothesize that this is due to an ``over-refinement" phenomenon. After several iterations, the intermediate result is already quite detailed. Excessive additional steps may cause the model to amplify minor noise or artifacts from the initial prediction, or even to hallucinate spurious textures that are plausible according to its generative prior but not faithful to the source image. Therefore, T=6 represents the optimal trade-off between performance and efficiency.

\subsubsection{Effectiveness of Diffusion Refinement.}

Finally, to validate the effectiveness of the diffusion module in our framework, we compared the full model against the results from the initial prediction module alone (ViT Module Only). The data reveals that after diffusion refinement, the LPIPS score significantly decreased by 40.98\%, 36.78\%, and 33.94\% on the J\&K, SA, and Med datasets, respectively. Meanwhile, it can also be seen from the \cref{compare7} that $D^3$-RSMDE outputs more texture information than ViT Module. These evidences strongly prove that our lightweight diffusion module dramatically enhances the perceptual quality and visual clarity of the generated depth maps.

\section{Conclusion}

In conclusion, even though inferring depth information from a single remote sensing image remains difficult for both models and human eyes, our comprehensive experiments strongly demonstrate the effectiveness and exceptional efficiency of the proposed D³-RSMDE framework. Our model achieves accuracy comparable to, and in some cases surpassing, the SOTA Marigold, a model renowned for its high-fidelity synthesis. More significantly, D³-RSMDE represents a breakthrough in computational efficiency. It uses PLBR strategy that provides a substantially accelerated training process and achieves an inference speedup of up to 40 times, all while maintaining a VRAM footprint on par with lightweight ViT-based architectures. Therefore, our work successfully addresses the critical trade-off between accuracy and efficiency, resolving the prohibitive computational bottleneck that has hindered the practical deployment of high-fidelity diffusion-based models in the field of RSMDE.

\section{Acknowledgments}
This work was sponsored by National Natural Science Foundation of China (62576305), and the Fundamental Research Funds for the Central Universities (No. 226-2025-00057).

\bibliography{aaai2026}

\end{document}